\documentclass[conference]{IEEEtran}
\IEEEoverridecommandlockouts
\usepackage{cite}
\usepackage{amsmath,amssymb,amsfonts}
\usepackage{algorithmic}
\usepackage{textcomp}
\usepackage{threeparttable}
\usepackage{xcolor}
\usepackage{url}
\usepackage{array}
\usepackage{booktabs,multirow,graphicx}
\usepackage{placeins}
\usepackage{float}
\usepackage{hyperref}
\usepackage{stfloats}

\def\BibTeX{{\rm B\kern-.05em{\sc i\kern-.025em b}\kern-.08em
    T\kern-.1667em\lower.7ex\hbox{E}\kern-.125emX}}
\begin{document}

\title{WaveSFNet: A Wavelet-Based Codec and Spatial--Frequency Dual-Domain Gating Network for Spatiotemporal Prediction}

\author{
    \IEEEauthorblockN{Xinyong Cai$^{1}$, Runming Xie$^{1}$, Hu Chen$^{2, 3}$, Yuankai Wu$^{2, 3, \ast}$\thanks{$^{\ast}$Corresponding author.}}
    \IEEEauthorblockA{$^{1}$College of Software Engineering, Sichuan University \quad $^{2}$College of Computer Science, Sichuan University}
    \IEEEauthorblockA{$^{3}$Sichuan University National Key Laboratory of Fundamental Science on Synthetic Vision \\
    fenghuajuedai09@gmail.com, A372707325@126.com, huchen@scu.edu.cn, wuyk0@scu.edu.cn}
}

\maketitle

\begin{abstract}
Spatiotemporal predictive learning aims to forecast future frames from historical observations in an unsupervised manner, and is critical to a wide range of applications. The key challenge is to model long-range dynamics while preserving high-frequency details for sharp multi-step predictions. Existing efficient recurrent-free frameworks typically rely on strided convolutions or pooling for sampling, which tends to discard textures and boundaries, while purely spatial operators often struggle to balance local interactions with global propagation. To address these issues, we propose WaveSFNet, an efficient framework that unifies a wavelet-based codec with a spatial--frequency dual-domain gated spatiotemporal translator. The wavelet-based codec preserves high-frequency subband cues during downsampling and reconstruction. Meanwhile, the translator first injects adjacent-frame differences to explicitly enhance dynamic information, and then performs dual-domain gated fusion between large-kernel spatial local modeling and frequency-domain global modulation, together with gated channel interaction for cross-channel feature exchange. Extensive experiments demonstrate that WaveSFNet achieves competitive prediction accuracy on Moving MNIST, TaxiBJ, and WeatherBench, while maintaining low computational complexity. Our code is available at \url{https://github.com/fhjdqaq/WaveSFNet}.
\end{abstract}


\section{Introduction}

Spatiotemporal predictive learning aims to generate future sequences based on historical observations and is pivotal for applications ranging from weather forecasting~\cite{shi2015convolutional,reichstein2019deep} and traffic flow prediction~\cite{zhang2017deep,fang2019gstnet,cheng2024rethinking} to autonomous driving~\cite{bhattacharyya2018long,kwon2019predicting}. A core challenge lies in the complex entanglement of multi-scale dynamics, where slowly evolving global trends coexist with abrupt local transients. Consequently, an effective model must capture long-range dependencies across the spatiotemporal field while rigorously preserving high-frequency details such as edges and textures to prevent error accumulation and blurring during multi-step prediction.

Existing approaches predominantly fall into recurrent-based and recurrent-free paradigms. Recurrent-based models, represented by ConvLSTM~\cite{shi2015convolutional} and the PredRNN family~\cite{wang2017predrnn,wang2018predrnn++,wang2022predrnn}, model temporal evolution via recursive state updates. Their step-by-step execution limits parallelization, resulting in high computational costs for high-resolution data. Conversely, recurrent-free methods~\cite{gao2022simvp,tan2023temporal} formulate prediction as an end-to-end sequence-to-sequence mapping. These models typically employ an encoder--translator--decoder pipeline, offering superior parallel efficiency and scalability. However, most recurrent-free architectures rely on standard strided convolutions or pooling for spatial sampling. Without explicit constraints on frequency preservation, this process often causes high-frequency information loss, manifesting as blurred textures and degraded details, particularly in dynamic scenes.

\begin{figure}[t]
    \centering
    \includegraphics[width=0.9\linewidth]{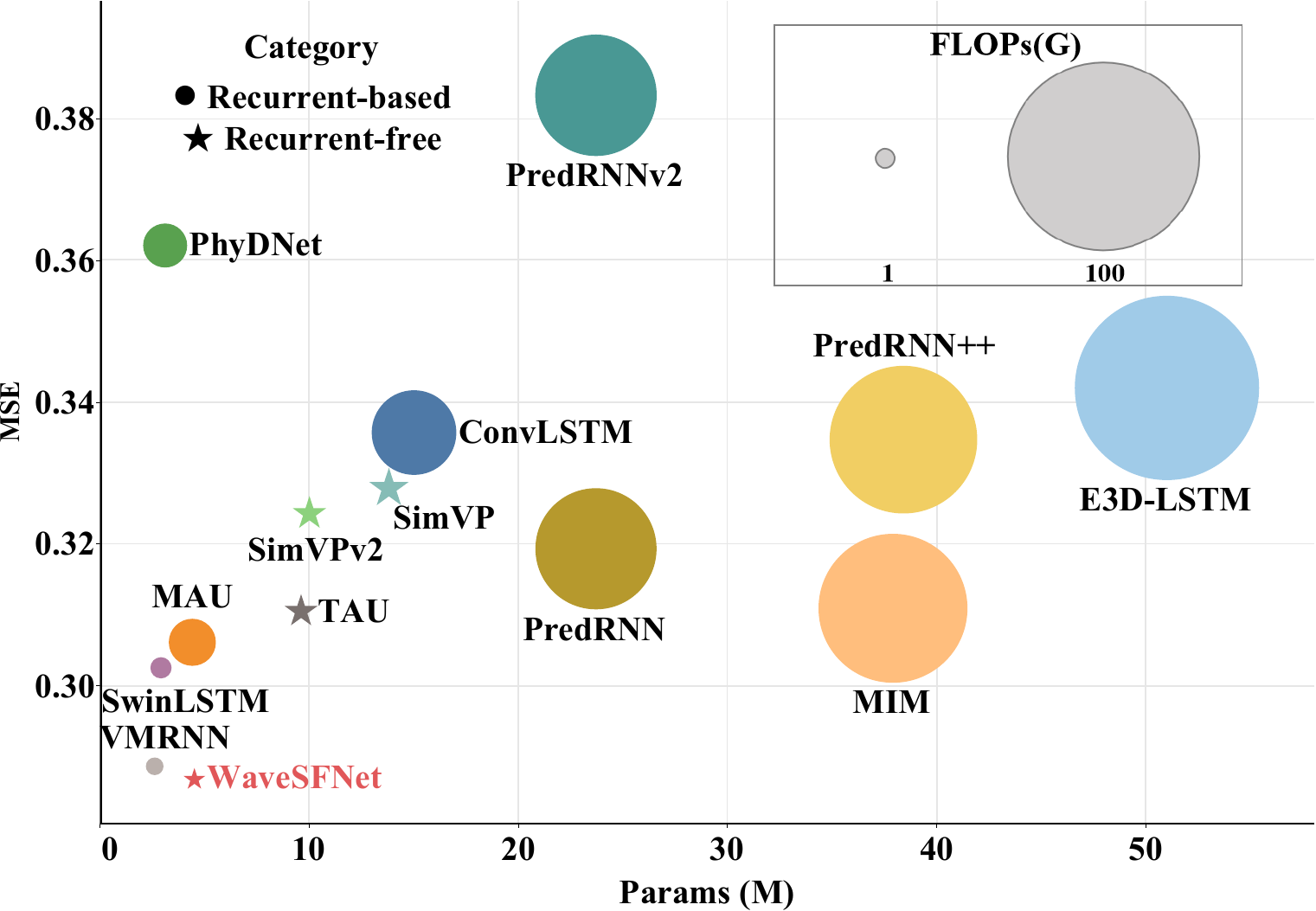}
    \vspace{-3pt}
    \caption{Performance comparison on the TaxiBJ dataset. Bubble size denotes FLOPs. WaveSFNet achieves the lowest MSE with reduced complexity.}
    \label{fig:teaser}
    \vspace{-15pt}
\end{figure}

From the perspective of representation learning, purely spatial operators often struggle to balance global propagation with local detail retention~\cite{wang2018non,dosovitskiy2021an}. Frequency-domain modeling enables efficient global modulation for capturing long-range dependencies~\cite{rao2021global,lee2022fnet}. Thus, a dual-domain strategy that synergizes spatial locality with spectral globality becomes essential. Complementing this, wavelet transforms provide a mathematically invertible framework that minimizes information loss during sampling by explicitly preserving high-frequency subbands~\cite{liu2018multi,li2020wavelet}. Therefore, it is promising to integrate dual-domain context modeling and multi-scale detail preservation within an efficient recurrent-free paradigm.

Motivated by these insights, we propose WaveSFNet, a wavelet-based spatial--frequency dual-domain gating network for spatiotemporal prediction. WaveSFNet adopts a streamlined encoder--translator--decoder architecture. The encoder leverages discrete wavelet transform for multi-scale downsampling and subband fusion, generating compact representations while preserving high-frequency cues. The translator stacks ST Blocks performing spatial--frequency dual-domain context extraction, fused via gated channel interaction. Finally, the decoder reconstructs the sequence through inverse discrete wavelet transform and a shallow skip connection. As highlighted in Fig.~\ref{fig:teaser}, this design achieves a superior trade-off between precision and efficiency on the TaxiBJ dataset, outperforming state-of-the-art baselines without relying on heavy attention mechanisms or recurrent inference.

Our main contributions are as follows:
\begin{itemize}
    \item We propose WaveSFNet, which replaces conventional convolutional encoders and decoders with wavelet-based codecs to better preserve structural details in spatiotemporal prediction.
    \item We design a dual-domain gated spatiotemporal translator, which simultaneously models local dynamics and global propagation in the latent space via adaptive channel-wise fusion.
    \item Extensive experiments across diverse domains demonstrate that our method achieves state-of-the-art performance with a favorable balance between prediction quality and computational cost.
\end{itemize}

\section{Related Work}

\subsection{Spatiotemporal Predictive Learning}
Existing spatiotemporal predictive learning methods are commonly grouped into recurrent-based and recurrent-free architectures.

Early recurrent approaches mainly combine recurrent neural networks with convolutional operators. ConvLSTM~\cite{shi2015convolutional} introduces convolutions into gated recurrent units, laying the foundation for spatiotemporal modeling. The PredRNN~\cite{wang2017predrnn} family strengthens spatiotemporal memory to better capture short-term dynamics and reduce error accumulation. Later, MIM~\cite{wang2019memory} enhances change modeling via differential memory, while PhyDNet~\cite{guen2020disentangling} incorporates physical priors to disentangle dynamics. More recently, SwinLSTM~\cite{tang2023swinlstm} adopts window-based attention to expand spatial dependency modeling, and VMRNN~\cite{tang2024vmrnn} introduces state-space modeling to improve long-sequence efficiency. Despite their strengths, recurrent models rely on step-by-step updates, limiting parallelism and often increasing training and inference costs.

Recurrent-free architectures remove the sequential bottleneck by formulating prediction as an end-to-end sequence-to-sequence mapping. SimVP~\cite{gao2022simvp} establishes a strong baseline with a concise encoder--translator--decoder pipeline, where the translator can be replaced by MetaFormer-style~\cite{yu2022metaformer,tan2023openstl} modular backbones to compare interaction mechanisms. TAU~\cite{tan2023temporal} decomposes temporal modeling for parallel motion capture. Earthformer~\cite{gao2022earthformer} targets high-resolution Earth system forecasting with scalable cuboid attention. MMVP~\cite{zhong2023mmvp} adopts a dual-stream design to separate motion and appearance. TAT~\cite{nie2024triplet} introduces triplet attention that factorizes interactions over temporal, spatial, and channel dimensions. While recurrent-free methods are highly parallelizable, purely spatial designs or fixed receptive fields may degrade details in highly dynamic scenes, especially high-frequency content.

\subsection{Wavelets and Frequency-Domain Learning}
Spectral methods use Fourier transforms for efficient global frequency modulation. GFNet~\cite{rao2021global} introduces learnable frequency-domain filters as an efficient alternative to expensive attention. FourCastNet~\cite{pathak2022fourcastnet} uses adaptive Fourier neural operators for global mixing in Earth system prediction. PastNet~\cite{wu2024pastnet} incorporates frequency priors into spatiotemporal forecasting, and SpectFormer~\cite{patro2025spectformer} combines spectral gating with attention to balance global modulation and local interactions. Standard Fourier representations emphasize global spectral modeling, and the incorporation of spatial locality can be considered for non-stationary signals with strong transients.

Wavelet-based approaches exploit multi-resolution analysis with spatial–frequency locality, enabling detail-preserving sampling.
MWCNN~\cite{liu2018multi} replaces conventional sampling with wavelet transforms to enlarge the receptive field while reducing information loss.
WaveViT~\cite{yao2022wave} integrates wavelet-based sampling into vision transformers for unified multi-scale learning.
For spatiotemporal prediction, WaST~\cite{nie2024wavelet} leverages multi-scale subbands to retain fine-grained structures during temporal forecasting.
Compared with standard sampling schemes, wavelet-based models preserve local details across scales and provide a faithful basis for subsequent modeling.

\section{Method}

\begin{figure*}[!t]
    \centering
    \includegraphics[width=0.95\linewidth]{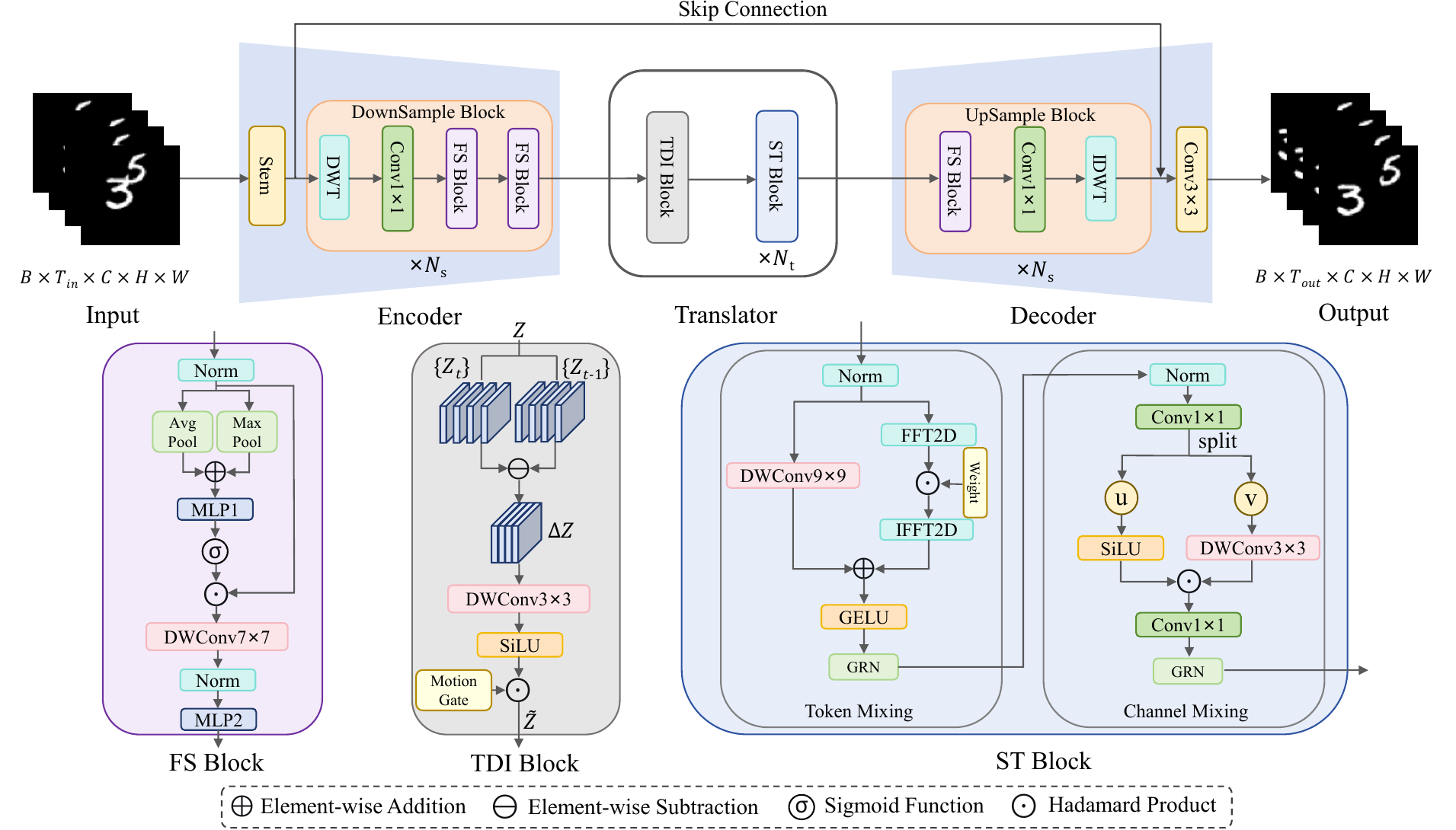}
    \vspace{-2pt}
    \caption{\textbf{Overall architecture and core modules of WaveSFNet.} WaveSFNet follows an encoder--translator--decoder design. A wavelet-based multi-scale encoder extracts latent features from input frames. A \textbf{TDI Block} injects adjacent-frame differences, and $N_t$ stacked \textbf{ST Blocks} apply spatial--frequency dual-domain gating after packing time into channels. A wavelet-symmetric decoder reconstructs predictions.}
    \label{fig:architecture1}
    \vspace{-7pt}
\end{figure*}

\subsection{Overall Framework}

Given an input spatiotemporal sequence $\mathbf{X}\in\mathbb{R}^{B\times T_{\mathrm{in}}\times C\times H\times W}$, we learn a parametric mapping $f_{\theta}$ to model the spatiotemporal dynamics and predict future frames:
\begin{equation}
\widehat{\mathbf{Y}}=f_{\theta}(\mathbf{X})\in\mathbb{R}^{B\times T_{\mathrm{out}}\times C\times H\times W}.
\end{equation}

As depicted in Fig.~\ref{fig:architecture1}, the overall framework consists of three components: 
(1) \textbf{Wavelet-based multi-scale encoder}, which decomposes each input frame into low-frequency structures and high-frequency details and fuses them into compact latent representations; 
(2) \textbf{Spatiotemporal translator}, which injects adjacent-frame difference cues into the coarse-scale latent space and extracts contextual information in parallel from spatial and frequency domains, followed by gated channel interaction; 
(3) \textbf{Wavelet-symmetric decoder}, which progressively reconstructs spatial resolution via inverse wavelet transforms and produces the predicted frames.

\subsection{Wavelet-based Multi-scale Encoding}

For each time step $t=1,\dots,T_{\mathrm{in}}$, the input frame is denoted as
$\mathbf{X}_t\in\mathbb{R}^{C\times H\times W}$. Shallow features are first extracted:

\begin{equation}
\mathbf{F}^{(0)}_t=
\operatorname{Conv}_{3\times3}\!\Big(
\operatorname{GELU}(\operatorname{Conv}_{3\times3}(\mathbf{X}_t))
\Big).
\end{equation}
The resulting shallow features satisfy
$\mathbf{F}^{(0)}_t \in \mathbb{R}^{C_0\times H\times W}$,
and are retained for high-resolution detail compensation in decoding.

Subsequently, $N_s$ wavelet downsampling stages are applied.
We employ the \textbf{Haar} wavelet for 2D DWT/IDWT, which is a \textbf{parameter-free and invertible} linear transform, enabling stable multi-scale decomposition and reconstruction.
At the $i$-th stage, a 2D discrete wavelet transform decomposes the features into four subbands:
\begin{equation}
\big(\mathbf{F}^{(i),LL}_t,\mathbf{F}^{(i),HL}_t,\mathbf{F}^{(i),LH}_t,\mathbf{F}^{(i),HH}_t\big)
=
\mathcal{W}\!\left(\mathbf{F}^{(i-1)}_t\right).
\end{equation}
The four subbands are concatenated along the channel dimension and projected via a pointwise convolution:
\begin{equation}
\mathbf{E}^{(i)}_t
=
\operatorname{Conv}_{1\times1}\!\left(
\operatorname{Concat}_{c}\left(
\{\mathbf{F}^{(i),b}_t\}_{b\in\mathcal{B}}
\right)
\right),
\end{equation}
where
$\mathbf{E}^{(i)}_t\in\mathbb{R}^{C_s\times H_i\times W_i}$, $\mathcal{B}=\{LL,HL,LH,HH\}$ and $H_i=H/2^i$, $W_i=W/2^i$.

This fusion compresses complementary subband information into a unified channel space, forming compact multi-scale representations.

Frequency-selective enhancement is then applied to $\mathbf{E}^{(i)}_t$ using two FS Blocks at each scale to obtain $\mathbf{F}^{(i)}_t$ for stable representation learning.
Specifically, normalized features are gated by global average and max pooled statistics, refined by large-kernel depthwise convolution, and mixed by pointwise channel interaction.

\subsection{Temporal Difference Injection}

To enhance sensitivity to rapid temporal variations, we inject adjacent-frame differences in the coarse latent space via a TDI (Temporal Difference Injection) Block.

We first project the last-scale features $\mathbf{F}^{(N_s)}_t$ into compact latent representations $\mathbf{Z}_t\in\mathbb{R}^{C_z\times H_{N_s}\times W_{N_s}}$ via a frame-wise pointwise convolution, and compute adjacent-frame differences $\Delta\mathbf{Z}_t$ (with $\Delta\mathbf{Z}_1=\mathbf{0}$). We then inject these differences by extracting local motion cues using a depthwise convolution and modulating them with channel-wise gating:
\begin{equation}
\widetilde{\mathbf{Z}}_t
=
\mathbf{Z}_t
+
\mathbf{g}\odot
\operatorname{SiLU}\!\Big(
\operatorname{DWConv}_{3\times3}(\Delta \mathbf{Z}_t)
\Big),
\end{equation}
where $\mathbf{g}$ denotes learnable gating coefficients broadcastable over spatial dimensions. This operation introduces explicit short-term dynamics in the coarse latent space, enhancing sensitivity to rapid variations.

\subsection{Dual-domain Gated Spatiotemporal Translation}

The injected sequence $\widetilde{\mathbf{Z}}=\{\widetilde{\mathbf{Z}}_t\}_{t=1}^{T_{\mathrm{in}}}$ is fed into a spatiotemporal translator composed of $N_t$ stacked ST Blocks.
We stack the temporal dimension into channels and denote the resulting channel-stacked features as
$\widetilde{\mathbf{Z}}^{\mathrm{p}}\in\mathbb{R}^{C_t\times H_{N_s}\times W_{N_s}}$,
where $C_t=T_{\mathrm{in}}C_z$.
Each ST Block consists of dual-domain context extraction and gated channel interaction.
The spatial branch emphasizes local neighborhood structures, while the frequency branch captures global patterns and long-range dependencies, and the two complement each other.

\subsubsection{Dual-domain Context Extraction}

The spatial branch employs a $9\times9$ depthwise convolution:
\begin{equation}
\mathbf{S}=\operatorname{DWConv}_{9\times9}(\widetilde{\mathbf{Z}}^{\mathrm{p}}).
\end{equation}
The frequency branch applies a 2D real Fourier transform, modulates the spectrum with learnable complex-valued weights, and transforms back to the spatial domain:
\begin{equation}
\mathbf{P}=
\mathcal{F}^{-1}\!\big(
\mathcal{F}(\widetilde{\mathbf{Z}}^{\mathrm{p}})\odot \boldsymbol{\Psi}
\big),
\end{equation}
where $\mathcal{F}(\cdot)$ and $\mathcal{F}^{-1}(\cdot)$ denote the 2D real Fourier transform and its inverse, respectively, and $\boldsymbol{\Psi}$ represents learnable complex weights.
As the frequency weights operate globally over the spectrum, this branch provides modulation with a global receptive field, which is well suited for modeling large-scale structure propagation and long-range dependencies.

We fuse the two branches as $\mathbf{U}=\mathbf{S}+\mathbf{P}$, and implement it with a residual calibration for stable training.

\begin{table*}[!b]
\vspace{-3pt}
\centering
\caption{Detailed experimental configurations for different datasets. $C_s$ denotes the base channel width of the wavelet codec, and $C_t$ denotes the time-packed channel width used by ST Blocks.}
\vspace{-0.5em}
\label{tab:implementation_details}
\setlength{\tabcolsep}{1.2mm}
\renewcommand{\arraystretch}{1.15}
\resizebox{0.9\linewidth}{!}{
\begin{tabular}{m{3.2cm} >{\centering\arraybackslash}m{1.4cm} c c c c c c c c c c}
\toprule
\multicolumn{2}{c}{Dataset} 
& $N_s$ & $N_t$ & $C_s$ & $C_t$ & $(C,H,W)$ & Batch Size 
& $T_{\mathrm{in}}$ & $T_{\mathrm{out}}$
& LR & Epochs \\
\midrule
\multicolumn{2}{c}{Moving MNIST} 
& 2 & 8 & 64 & 720 & $(1,64,64)$ & 16  
& 10 & 10
& $9.2\times10^{-4}$ & 2000 \\
\multicolumn{2}{c}{TaxiBJ}        
& 1 & 8 & 32 & 192 & $(2,32,32)$ & 16  
& 4 & 4
& $1.5\times10^{-3}$ & 50 \\
\midrule
\multirow{5}{*}{\hspace{28pt}WeatherBench} 
& \hspace{-28pt}T2M  
& 1 & 8 & 32 & 264 & $(1,32,64)$ & 16 
& 12 & 12
& $2\times10^{-3}$ & 50 \\
& \hspace{-28pt}TCC  
& 1 & 8 & 32 & 264 & $(1,32,64)$ & 16 
& 12 & 12
& $4\times10^{-3}$ & 50 \\
& \hspace{-28pt}UV10 
& 1 & 8 & 32 & 264 & $(2,32,64)$ & 16 
& 12 & 12
& $2\times10^{-3}$ & 50 \\
& \hspace{-28pt}R    
& 1 & 8 & 32 & 264 & $(1,32,64)$ & 16 
& 12 & 12
& $3\times10^{-3}$ & 50 \\
& \hspace{-28pt}MV    
& 1 & 8 & 32 & 264 & $(12,32,64)$ & 16 
& 4 & 4
& $3\times10^{-3}$ & 50 \\
\bottomrule
\end{tabular}}
\vspace{-5pt}
\end{table*}

\subsubsection{Gated Channel Interaction}

Gated channel interaction is applied to enable adaptive cross-channel information exchange.
We first use a pointwise convolution to expand the contextual features, and then split the result along the channel dimension into two parts,
$\mathbf{U}^{(u)}$ and $\mathbf{U}^{(v)}$,
where $\mathbf{U}^{(u)}$ provides gating signals and $\mathbf{U}^{(v)}$ carries feature content.

A $3\times3$ depthwise convolution is applied to the feature branch, followed by element-wise modulation with a SiLU gate and projection back to the original channel dimension:
\begin{equation}
\mathbf{G}=
\operatorname{Conv}_{1\times1}\!\Big(
\operatorname{SiLU}(\mathbf{U}^{(u)})\odot
\operatorname{DWConv}_{3\times3}(\mathbf{U}^{(v)})
\Big).
\end{equation}

This gated interaction enables nonlinear channel mixing, facilitating the exchange of frequency-aware semantics and temporal cues across channels.
GRN~\cite{woo2023convnext}, layer scale~\cite{touvron2021going}, and DropPath are applied throughout the spatiotemporal translator to stabilize training and provide regularization.
After stacking $N_t$ ST Blocks, we reshape the output to unstack time from channels, followed by a frame-wise pointwise convolution that maps $C_z$ back to $C_s$, producing $\{\mathbf{Q}_t\}_{t=1}^{T_{\mathrm{in}}}\in\mathbb{R}^{T_{\mathrm{in}}\times C_s\times H_{N_s}\times W_{N_s}}$.

\subsection{Wavelet-symmetric Decoding}
 
The decoder takes the translated latent sequence $\{\mathbf{Q}_t\}_{t=1}^{T_{\mathrm{in}}}$ as input and mirrors the encoder with $N_s$ upsampling stages.

At each stage, the latent features are refined by an FS Block to better support inverse wavelet reconstruction, followed by a pointwise convolution and a 2D inverse wavelet transform to restore the spatial resolution from $(H_i,W_i)$ to $(H_{i-1},W_{i-1})$.

After the final stage, we fuse the shallow features $\mathbf{F}^{(0)}_t$ from the encoder via a skip connection to recover fine-grained details. A $3\times3$ readout convolution is then applied to produce the predicted frames $\widehat{\mathbf{Y}}_t\in\mathbb{R}^{C\times H\times W}$, $t=1,\dots,T_{\mathrm{out}}$, and the outputs are stacked over time to form $\widehat{\mathbf{Y}}$.

By default, $T_{\mathrm{out}} = T_{\mathrm{in}}$.  
When $T_{\mathrm{out}} < T_{\mathrm{in}}$, the first $T_{\mathrm{out}}$ frames of the predicted sequence are retained;  
when $T_{\mathrm{out}} > T_{\mathrm{in}}$, an autoregressive rollout strategy is adopted, where the previously predicted sequence is iteratively fed back as input until reaching length $T_{\mathrm{out}}$.

\section{Experiments}

\subsection{Experimental Setup}

\subsubsection{Datasets}
Following the OpenSTL benchmark protocol~\cite{tan2023openstl}, we evaluate our method on three datasets from distinct domains.

\textbf{Moving MNIST}~\cite{srivastava2015unsupervised} is a synthetic motion benchmark of two moving handwritten digits on $64\times64$ grids. We use 10,000 sequences for training and 10,000 for testing.
\textbf{TaxiBJ}~\cite{zhang2017deep} contains real-world Beijing traffic inflow and outflow fields on $32\times32$ grids with two channels. We use 20,461 samples for training and 500 for testing.
\textbf{WeatherBench}~\cite{rasp2020weatherbench} is a global weather forecasting benchmark with multiple spatial resolutions. 
We evaluate two settings on the $5.625^\circ$ version. 
(1) \emph{Single-variable} forecasting for temperature (T2M), humidity (R), wind components (UV10), and cloud cover (TCC), using data from 2010--2015 for training, 2016 for validation, and 2017--2018 for testing. 
(2) \emph{Multi-variable (MV)} forecasting jointly predicts humidity (R), temperature (T), longitude wind (U), and latitude wind (V) at 150, 500, and 850 hPa, using a longer training period (1979--2015) and the same validation (2016) and testing (2017--2018) years.

\subsubsection{Metrics}
We employ standard metrics for comprehensive evaluation. MSE and MAE are used to quantify prediction errors (lower is better), with RMSE additionally reported for weather tasks. SSIM and PSNR are used to assess visual quality and structural fidelity (higher is better). Furthermore, model efficiency is evaluated by the number of parameters (Params) and floating-point operations (FLOPs). 

\subsubsection{Implementation Details}
Our model is implemented using PyTorch within the OpenSTL framework~\cite{tan2023openstl}. We utilize the Adam optimizer for all experiments, minimizing the Mean Squared Error (MSE) loss function. The learning rate is scheduled via a OneCycle strategy for Moving MNIST and a Cosine Annealing strategy for the other datasets. The specific architectural hyperparameters and detailed training configurations are summarized in Table~\ref{tab:implementation_details}.

\subsection{Comparisons with State-of-the-Art}
\label{sec:sota_comparisons}
We compare our method with two categories: \textit{recurrent-based} approaches built on recurrent networks (e.g., ConvLSTM, PredRNN, and VMRNN), and \textit{recurrent-free} approaches without recurrence (e.g., SimVP, TAU, and MogaNet). 

\begin{table}[htbp]
\vspace*{-8pt}
\centering
\caption{Quantitative comparison of different methods on Moving MNIST.}
\vspace{-0.5em}
\begin{tabular}{c c c c}
\hline
Category & Method & MSE $\downarrow$ & SSIM $\uparrow$ \\
\hline
\multirow{9}{*}{Recurrent-based} & ConvLSTM~\cite{shi2015convolutional} & 103.3 & 0.707 \\
 & FRNN~\cite{oliu2018folded} & 69.7 & 0.813 \\
 & PredRNN~\cite{wang2017predrnn} & 56.8 & 0.867 \\
 & MIM~\cite{wang2019memory} & 44.2 & 0.910 \\
 & MAU~\cite{chang2021mau} & 27.6 & 0.937 \\
 & PhyDNet~\cite{guen2020disentangling} & 24.4 & 0.947 \\
 & CrevNet~\cite{yu2020efficient} & 22.3 & 0.949 \\
 & SwinLSTM~\cite{tang2023swinlstm} & 17.7 & 0.962 \\ 
 & VMRNN~\cite{tang2024vmrnn} & \underline{16.5} & \underline{0.965} \\  
\hline
\multirow{5}{*}{Recurrent-free} & SimVP~\cite{gao2022simvp} & 23.8 & 0.948 \\
 & MMVP~\cite{zhong2023mmvp} & 22.2 & 0.952 \\
 & TAU~\cite{tan2023temporal} & 19.8 & 0.957 \\
 & TAT~\cite{nie2024triplet} & 17.6 & 0.960 \\
 & Ours & \textbf{15.8} & \textbf{0.966} \\ 
\hline
\end{tabular}
\label{tab:mmnist_comparison}
\vspace{-2.15pt}
\end{table}

\noindent\textbf{Results on Synthetic Motion.}
Table~\ref{tab:mmnist_comparison} reports quantitative results on Moving MNIST. Our method achieves the best MSE (15.8) and SSIM (0.966). This superior structural fidelity stems from our wavelet-based codec explicitly preserving high-frequency edges, which effectively mitigates the texture blurring typically observed in long-horizon predictions.

\begin{figure}[htbp]
    \centering
    \includegraphics[width=0.985\linewidth]{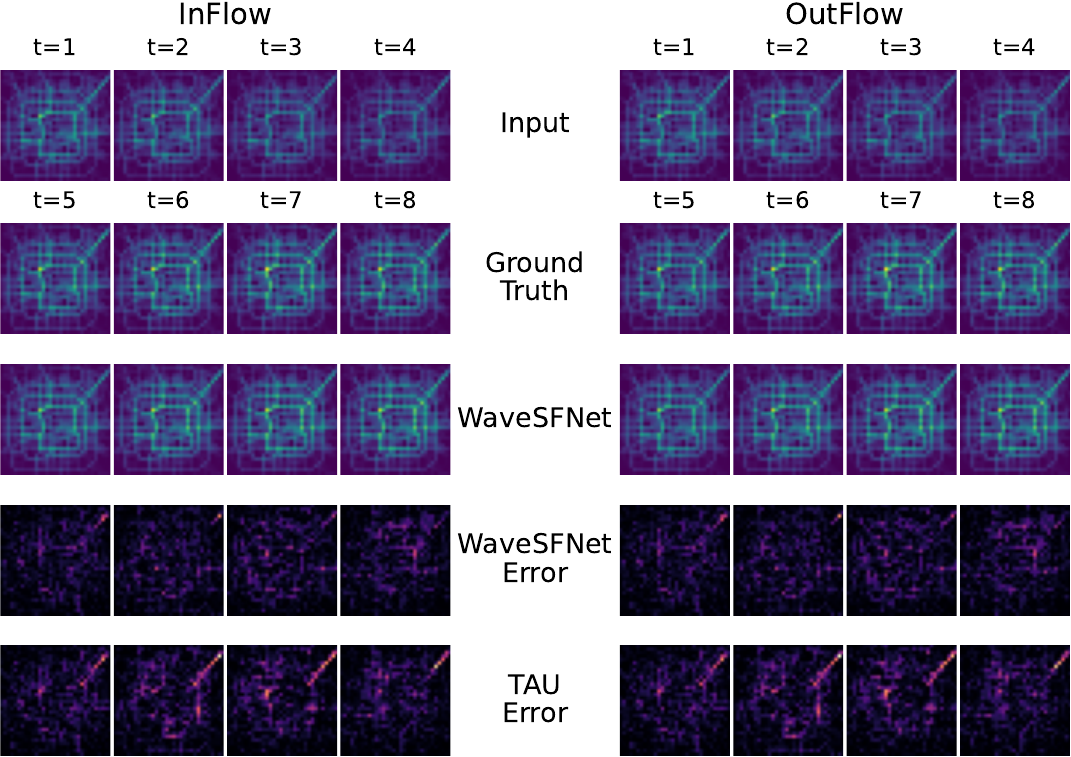}
    \vspace{-0.5pt}
    \caption{Qualitative visualizations of WaveSFNet on TaxiBJ.}
    \label{fig:taxibj_vis}
\end{figure}

\begin{table}[htbp]
\vspace*{-2pt}
\centering
\caption{Quantitative comparison of different methods on TaxiBJ.}
\vspace{-0.5em}
\setlength{\tabcolsep}{2.5pt} 
\renewcommand{\arraystretch}{1.07}
\scriptsize
\resizebox{\columnwidth}{!}{%
\begin{tabular}{c c c c c c c c}
\toprule
Category & Method & \shortstack{Params (M)} & \shortstack{FLOPs (G)}
& MSE$\downarrow$ & MAE$\downarrow$ & SSIM$\uparrow$ & PSNR$\uparrow$ \\
\midrule
\multirow{10}{*}{Recurrent-based} 
& ConvLSTM~\cite{shi2015convolutional}   & 15.0 & 20.7  & 0.3358 & 15.32 & 0.9836 & 39.45 \\
& PredRNN~\cite{wang2017predrnn}    & 23.7 & 42.4  & 0.3194 & 15.31 & 0.9838 & 39.51 \\
& PredRNN++~\cite{wang2018predrnn++}  & 38.4 & 63.0  & 0.3348 & 15.37 & 0.9834 & 39.47 \\
& E3D-LSTM~\cite{wang2018eidetic}   & 51.0 & 98.19 & 0.3421 & 14.98 & 0.9842 & 39.64 \\
& PhyDNet~\cite{guen2020disentangling}    & 3.1  & 5.6   & 0.3622 & 15.53 & 0.9828 & 39.46 \\
& MIM~\cite{wang2019memory}        & 37.9 & 64.1  & 0.3110 & 14.96 & 0.9847 & 39.65 \\
& MAU~\cite{chang2021mau}        & 4.4  & 6.4   & 0.3062 & 15.26 & 0.9840 & 39.52 \\
& PredRNNv2~\cite{wang2022predrnn}  & 23.7 & 42.6  & 0.3834 & 15.55 & 0.9826 & 39.49 \\
& SwinLSTM~\cite{tang2023swinlstm}   & \underline{2.9}  & 1.3   & 0.3026 & 15.00 & 0.9843 & -- \\
& VMRNN~\cite{tang2024vmrnn}      & \textbf{2.6}  & \textbf{0.9}   & \underline{0.2887} & \underline{14.69} & \underline{0.9858} & -- \\
\midrule
\multirow{5}{*}{Recurrent-free}
& SimVP~\cite{gao2022simvp}      & 13.8 & 3.6   & 0.3282 & 15.45 & 0.9835 & 39.45 \\
& SimVPv2~\cite{tan2211simvp}    & 10.0 & 2.6   & 0.3246 & 15.03 & 0.9844 & 39.71 \\
& TAU~\cite{tan2023temporal}        & 9.6  & 2.5   & 0.3108 & 14.93 & 0.9848 & \underline{39.74} \\
& MogaNet~\cite{moga}    & 10.0 & 2.6   & 0.3114 & 15.06 & 0.9847 & 39.70 \\
& Ours       & 4.5  & \underline{1.1}   & \textbf{0.2870} & \textbf{14.68} & \textbf{0.9859} & \textbf{39.88} \\
\bottomrule
\end{tabular}%
}
\label{tab:taxibj_comparison}
\vspace{-10pt}
\end{table}

\begin{table*}[htbp]
\centering
\caption{Quantitative comparison of different methods on WeatherBench single-variable tasks, including UV10, T2M, TCC, and R.}
\vspace{-0.5em}
\setlength{\tabcolsep}{1.5mm}
\renewcommand{\arraystretch}{1.08}
\resizebox{0.99\linewidth}{!}{
\begin{tabular}{c|cc|ccc|ccc|ccc|ccc}
\hline
\multirow[c]{2}{*}{Method} &
\multirow[c]{2}{*}{Params (M)} &
\multirow[c]{2}{*}{FLOPs (G)} &
\multicolumn{3}{c|}{UV10} &
\multicolumn{3}{c|}{T2M} &
\multicolumn{3}{c|}{TCC} &
\multicolumn{3}{c}{R} \\
&&&
MSE$\downarrow$ & MAE$\downarrow$ & RMSE$\downarrow$ &
MSE$\downarrow$ & MAE$\downarrow$ & RMSE$\downarrow$ &
MSE$\downarrow$ & MAE$\downarrow$ & RMSE$\downarrow$ &
MSE$\downarrow$ & MAE$\downarrow$ & RMSE$\downarrow$ \\
\hline
ConvLSTM~\cite{shi2015convolutional}   & 14.98 & 136   & 1.8976 & 0.9215 & 1.3775 & 1.5210 & 0.7949 & 1.2330 & 0.0494 & 0.1542 & 0.2223 & 35.1460 & 4.0120 & 5.9280 \\
PredRNN~\cite{wang2017predrnn}    & 23.57 & 278   & 1.8810 & 0.9068 & 1.3715 & 1.3310 & 0.7246 & 1.1540 & 0.0550 & 0.1588 & 0.2346 & 37.6110 & 4.0960 & 6.1330 \\
PredRNN++~\cite{wang2018predrnn++}  & 38.40 & 413   & 1.8727 & 0.9019 & 1.3685 & 1.4580 & 0.7676 & 1.2070 & 0.0548 & 0.1544 & 0.2341 & 45.9930 & 4.7310 & 6.7820 \\
MAU~\cite{chang2021mau}        & \underline{5.46} & 39.6  & 1.9001 & 0.9194 & 1.3784 & 1.2510 & 0.7036 & 1.1190 & 0.0496 & 0.1516 & 0.2226 & 34.5290 & 4.0040 & 5.8760 \\
SimVP~\cite{gao2022simvp}      & 14.67 & 8.03  & 1.9993 & 0.9510 & 1.4140 & 1.2380 & 0.7037 & 1.1130 & 0.0477 & 0.1503 & 0.2183 & 34.3550 & 3.9940 & 5.8610 \\
HorNet~\cite{rao2022hornet}     & 12.42 & 6.85  & \underline{1.5539} & \underline{0.8254} & \underline{1.2466} & 1.2010 & 0.6906 & 1.0960 & \underline{0.0469} & 0.1475 & 0.2166 & 32.0810 & 3.8260 & 5.6640 \\
TAU~\cite{tan2023temporal}        & 12.22 & 6.70  & 1.5925 & 0.8426 & 1.2619 & 1.1620 & 0.6707 & 1.0780 & 0.0472 & 0.1460 & 0.2173 & 31.8310 & 3.8180 & 5.6420 \\
MogaNet~\cite{moga}    & 12.76 & 7.01  & 1.6072 & 0.8451 & 1.2678 & 1.1520 & 0.6665 & 1.0730 & 0.0470 & 0.1480 & 0.2168 & \underline{31.7950} & 3.8160 & 5.6390 \\
WaST~\cite{nie2024wavelet}       & \textbf{3.90} & \textbf{1.50}
           & -      & -      & -
           & \underline{1.0980} & \underline{0.6338} & \underline{1.0440}
           & -      & \textbf{0.1452} & \underline{0.2150}
           & -      & \textbf{3.6940} & \underline{5.5690} \\
Ours       & 8.73 & \underline{4.06}
           & \textbf{1.4601} & \textbf{0.8063} & \textbf{1.2084}
           & \textbf{1.0458} & \textbf{0.6311} & \textbf{1.0226}
           & \textbf{0.0459} & \underline{0.1459} & \textbf{0.2143}
           & \textbf{30.4813} & \underline{3.7363} & \textbf{5.5210} \\
\hline
\end{tabular}
}
\label{table:comp1}
\vspace{-10pt}
\end{table*}

\noindent\textbf{Results on Urban Traffic Flow.}
Urban traffic flow forecasting requires capturing both nonlinear spatiotemporal dependencies and periodic variations. As shown in Table~\ref{tab:taxibj_comparison}, our method consistently leads in all metrics. Qualitative comparisons in Fig.~\ref{fig:taxibj_vis} further validate this superiority; the error heatmaps demonstrate that WaveSFNet yields significantly lower residuals than TAU, indicating a more precise capture of complex flow dynamics. Crucially, we achieve this with highly competitive computational cost (4.5M Params, 1.1G FLOPs). This optimal accuracy--efficiency trade-off makes our method suitable for deployment on resource-constrained edge devices in intelligent transportation systems.

\begin{figure}[htbp]
    \centering
    \includegraphics[width=0.65\linewidth]{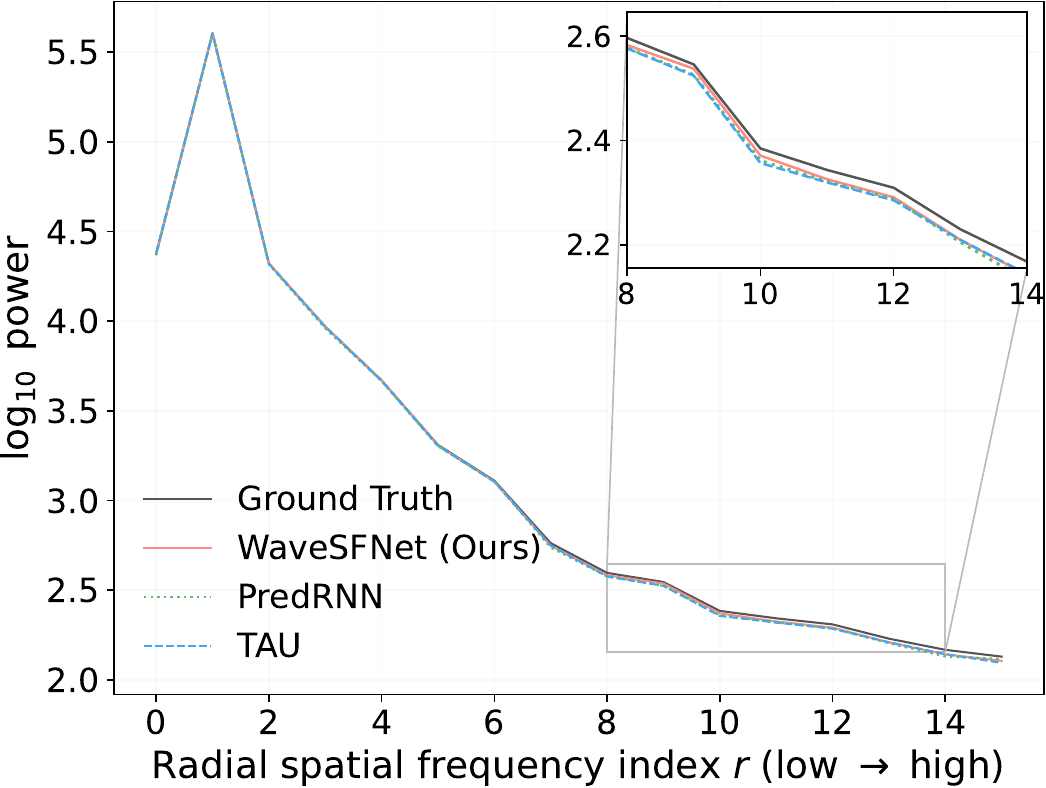}
    \vspace{-0.5pt}
    \caption{Frequency spectrum analysis on the WeatherBench T2M dataset. The plot displays the Radially Averaged Power Spectral Density, where the x-axis represents the radial spatial frequency and the y-axis denotes the log power spectrum. The inset provides a zoomed-in view of the high-frequency region.}
    \label{fig:frequency}
    \vspace{-3pt}
\end{figure}

\begin{table}[htbp]
\centering
\caption{Quantitative comparison of different methods on WeatherBench multi-variable forecasting (R, T, U, and V). All metrics are averaged over the four variables.}
\vspace{-0.5em}

\setlength{\tabcolsep}{2.0mm}
\renewcommand{\arraystretch}{1.08}
\footnotesize

\resizebox{0.99\linewidth}{!}{
\begin{tabular}{cccccc}
\toprule
Method & Params (M) & FLOPs (G) & MSE$\downarrow$ & MAE$\downarrow$ & RMSE$\downarrow$ \\
\midrule
ConvLSTM~\cite{shi2015convolutional}         & 15.50            & 43.33            & 108.81             & 5.7439             & 8.1810 \\
PredRNN~\cite{wang2017predrnn}               & 24.56            & 88.02            & 104.16             & 5.5373             & 7.9553 \\
PredRNN++~\cite{wang2018predrnn++}           & 39.31            & 129.0            & 106.77             & 5.5821             & 8.0568 \\
MIM~\cite{wang2019memory}                    & 41.71            & 35.77            & 121.95             & 6.2786             & 8.7376 \\
MAU~\cite{chang2021mau}                      & \underline{5.46}             & 12.07            & 106.13             & 5.6487             & 7.9928 \\
PredRNNv2~\cite{wang2022predrnn}             & 24.58            & 88.49            & 108.94             & 5.7747             & 8.1872 \\
SimVP~\cite{gao2022simvp}                    & 13.80            & 7.26             & 108.50             & 5.7360             & 8.1165 \\
SimVPv2~\cite{tan2211simvp}                  & 9.96             & 5.25             & 103.36             & 5.4856             & 7.9059 \\
ConvMixer~\cite{trockman2023patches}         & \textbf{0.85}    & \textbf{0.49}    & 112.76             & 5.9114             & 8.3238 \\
HorNet~\cite{rao2022hornet}                  & 9.68             & 5.12             & 104.21             & 5.5181             & 7.9854 \\
MogaNet~\cite{moga}                          & 9.97             & 5.25             & \underline{98.664} & \underline{5.3003} & \underline{7.6539} \\
TAU~\cite{tan2023temporal}                   & 9.55             & 5.01             & 99.428             & 5.3282             & 7.6855 \\
Ours                                         & 8.26             & \underline{3.73} & \textbf{96.141}    & \textbf{5.2550}    & \textbf{7.5569} \\
\bottomrule
\end{tabular}
}
\label{table:mv_bench_mean}
\vspace{-10pt}
\end{table}

\noindent\textbf{Results on Weather Forecasting.}
Weather forecasting is challenging due to chaotic atmospheric dynamics and complex multi-scale dependencies. For single-variable forecasting (Table~\ref{table:comp1}), our method remains robust across diverse variables. This trend aligns with our motivation: weather fields exhibit smooth structures and high-frequency variations, where dual-domain modeling can be beneficial.

To assess structural fidelity, we conduct a spectral analysis on the T2M dataset by computing the Radially Averaged Power Spectral Density (RAPSD) on the last predicted frame of all 17,495 test sequences. As shown in Fig.~\ref{fig:frequency}, accurately capturing high-frequency components remains a formidable challenge, with all models deviating from the Ground Truth. PredRNN and TAU exhibit a more pronounced high-frequency power decay, while WaveSFNet demonstrates a closer spectral alignment, suggesting that our overall architecture better mitigates the smoothing bias.

For multi-variable forecasting (Table~\ref{table:mv_bench_mean}), our method achieves the best overall accuracy with only 3.73G FLOPs, suggesting that WaveSFNet can capture cross-variable dependencies under limited computational budgets, scaling elegantly across diverse meteorological factors.

\subsection{Ablation Study}

We perform ablation studies under the standard protocol on TaxiBJ and WeatherBench (TCC). All variants share the same training pipeline; only the specified component is modified.

\begin{table}[htbp]
\centering
\vspace{-4pt}
\caption{Ablation results on TaxiBJ and TCC.}
\vspace{-0.5em}
\label{tab:ablation}
\setlength{\tabcolsep}{4pt}
\renewcommand{\arraystretch}{1.35}
\scriptsize
\begin{tabular}{c|ccc|cc}
\hline
\multirow{2}{*}{Variant} & \multicolumn{3}{c|}{TaxiBJ} & \multicolumn{2}{c}{TCC} \\
& MSE$\downarrow$ & MAE$\downarrow$ & SSIM$\uparrow$ & MSE$\downarrow$ & MAE$\downarrow$ \\
\hline
Spatial-only    & 0.2949 & 14.76 & 0.9854 & 0.04604 & 0.1494 \\
Frequency-only  & 0.3014 & 14.72 & 0.9850 & 0.04653 & 0.1518 \\
w/o TDI Block   & 0.2962 & 14.76 & 0.9854 & 0.04598 & 0.1489 \\ 
Conv codec      & 0.3198 & 14.92 & 0.9848 & 0.04609 & 0.1472 \\
MLP mixer       & 0.2997 & 14.91 & 0.9850 & 0.04620 & 0.1476 \\
WaveSFNet          & \textbf{0.2870} & \textbf{14.68} & \textbf{0.9859} & \textbf{0.04593} & \textbf{0.1459} \\
\hline
\end{tabular}
\vspace{-0.1pt}
\end{table}

As summarized in Table~\ref{tab:ablation}, we evaluate five variants: \emph{Spatial-only} removes the frequency branch in the dual-domain context extraction; \emph{Frequency-only} removes the spatial branch; \emph{w/o TDI Block} removes the TDI Block; \emph{Conv codec} replaces the wavelet-based codec with a convolutional one; \emph{MLP mixer} replaces gated channel interaction with standard MLP mixing.

Table~\ref{tab:ablation} shows that the full model achieves the best overall performance across both datasets. Removing either branch degrades accuracy, highlighting the complementarity between local spatial modeling and global spectral modulation. The performance drop observed in the \emph{w/o TDI Block} variant suggests that explicitly injecting short-term motion cues enhances the model's sensitivity to rapid temporal dynamics. Replacing the wavelet codec leads to the largest drop, suggesting that wavelet-based multi-scale decomposition mitigates information loss during sampling and supports more faithful reconstruction. Replacing multiplicative gating with MLP mixing also harms performance, indicating the benefit of gated channel interaction for feature exchange.

\section{Conclusion}
In this work, we propose WaveSFNet, an efficient recurrent-free spatiotemporal prediction architecture. By integrating a wavelet codec with a spatial--frequency dual-domain gating mechanism, WaveSFNet effectively alleviates high-frequency information loss during sampling while jointly modeling local dynamic variations and global long-range dependencies. Extensive experiments on synthetic motion, urban traffic flow, and global weather forecasting benchmarks demonstrate that WaveSFNet achieves state-of-the-art performance with a favorable balance between accuracy and computational cost. These results validate the effectiveness of incorporating multi-resolution spectral priors for high-fidelity spatiotemporal modeling, and suggest a promising direction for learning complex dynamic systems.

\section*{Acknowledgment}
This work was supported by the National Natural Science Foundation of China (No. 62406206), the Fundamental Research Funds for the Central Universities, and the Sichuan Provincial Natural Science Foundation (No. 2026NSFSC0426). We also appreciate the foundational work of pioneers in this field.

\FloatBarrier
\bibliographystyle{IEEEtran}
\bibliography{reference}

\end{document}